\documentclass[a4paper]{article}
\usepackage{graphicx}
\usepackage{cite}
\usepackage{multirow,booktabs}
\usepackage{caption}
\usepackage{subfigure}
\newcommand{\minitab}[2][l]{\begin{tabular}{#1}#2\end{tabular}}

\title{Learning More Robust Features with Adversarial Training}
\author{Shuangtao Li\footnote{Department of Computer Science and Technology, Chengdu University of Technology, Chengdu, Sichuan, China \newline E-mail: 724202992@qq.com} \and Yuanke Chen\footnote{Department of Geological Engineering, Chengdu University of Technology, Chengdu, Sichuan, China \newline E-mail: 1015575076@qq.com} \and Yanlin Peng\footnote{Department of Information and Computing Sciences, Chengdu University of Technology, Chengdu, Sichuan, China \newline E-mail: 1047950416@qq.com} \and Lin Bai\footnote{Geomathematics Key Laboratory of Sichuan Province, Chengdu, Sichuan, China \newline E-mail: bailin@mail.cdut.edu.cn}}

\date{}

\begin{document}
\maketitle

\begin{abstract}
In recent years, it has been found that neural networks can be easily fooled by adversarial examples, which is a potential safety hazard in some safety-critical applications. Many researchers have proposed various method to make neural networks more robust to white-box adversarial attacks, but an effective method have not been found so far. In this short paper, we focus on the robustness of the features learned by neural networks. We show that the features learned by neural networks are not robust, and find that the robustness of the learned features is closely related to the resistance against adversarial examples of neural networks. We also find that adversarial training against fast gradients sign method (FGSM) does not make the leaned features very robust, even if it can make the trained networks very resistant to FGSM attack. Then we propose a method, which can be seen as an extension of adversarial training, to train neural networks to learn more robust features. We perform experiments on MNIST and CIFAR-10 to evaluate our method, and the experiment results show that this method greatly improves the robustness of neural networks and the features they learn.
\end{abstract}
\textbf{\large Keywords:} Adversarial examples, Adversarial training, Neural networks, Robustness
\section{Introduction}
Although neural networks have been applied in a variety of tasks, they have proven to be vulnerable to adversarial examples \cite{ref1}, the examples slightly perturbed using some attack technique such as Fast Gradient Sign Method (FGSM) \cite{ref2} or C\&W's Attack \cite{ref3}. To defend adversarial examples, researchers have proposed many methods to increase the robustness of machine learning models, and the most famous one is adversarial training.

The simplest version of adversarial training proposed in \cite{ref2} trains models against the examples generated by FGSM, and the models trained by it are much more resistant to FGSM but not significantly more resistant to stronger adversaries \cite{ref6}, e.g. C\&W's Attack. \cite{ref4} proposed to train models against a stronger adversary projected gradient descent (PGD) attack. They found that adversarial training against PGD attack can significantly increase the resistance to adversarial examples generated by various techniques significantly, including PGD and C\&W’s Attack. However, their method requires computing the gradients w.r.t. the inputs dozens of times to generate adversarial examples to update the parameters of a model once. \cite{ref8} found "error amplification effect" that small adversarial perturbations added to input will be amplified in high-level representations, which cause neural networks to make wrong predictions. 

In this paper, to study how adversarial training influences neural networks, we focus on the robustness of the features learned by neural networks. We evaluate the robustness of the features by computing the distortions of the normalized values of the features, and we show that the features learned by neural networks are not robust. We find that adversarial training against FGSM can make the learned features more robust but still not very robust, even if it can make the trained networks very resistant to FGSM attack. To train neural networks to learn more robust features, we propose a method to make the learned features be distorted less by the perturbations added to the inputs.

In the rest of this paper, we first briefly introduce FGSM, PGD attack and adversarial training in Section 2. In Section 3, we first study the robustness of the features learned by neural networks, and then we detail our simple method for learning more robust features. The proposed method will be empirically evaluated in Section 4. Finally we draw a conclusion in Section 5.
\section{Background}
\textbf{Fast gradient sign method (FGSM).} Fast gradient sign method is a computationally efficient method for generating adversarial examples. It perturbs an example in the direction of the sign of the gradient of the loss function w.r.t. the input:
	
\begin{equation}
x^{*}=x+\epsilon \cdot sign(\nabla_x \cdot J(\theta,x,y))
\end{equation}
	
where $x$ is an example, $y$ is the label of $x$, $J_{(\theta, x, y)}$ is the loss function used to generate adversarial examples, $x^{*}$ is the perturbed example and $\epsilon$ is a constant used to control the size of perturbations. Though FGSM attack is computationally efficient, its success rate is usually relatively low.
	
\textbf{Projected gradient descent (PGD) attack.} Projected gradient descent attack generates adversarial examples by iteratively applying FGSM and projecting the perturbed example to be a valid example multiple times. It is a stronger adversary than FGSM, i.e. it can attack models with higher success rates.
	
\textbf{Adversarial training.} Adversarial training trains a model with this objective function:
	
\begin{equation}
J_{adv}(\theta, x, y) = \alpha\cdot J(\theta, x, y)+(1-\alpha)\cdot J(\theta, x^{*}, y),
\end{equation}
	
where $\alpha$ is a constant used to control the strength of the adversary, and $x^{*}$ can be generated by any method. In this paper we use FGSM to generate adversarial examples in adversarial training.
	
\section{Learning more robust features}
In this section, we first introduce how we evaluate the robustness of the features learned by neural networks and make observations about the robustness of the learned features. Then we detail our simple method to train neural networks to learn more robust features.
	
\subsection{Robustness of Learned Features}
Perturbations to inputs can cause distortions of the values of the features in the hidden layers, and the values of more robust features will be distorted less by adversarial perturbations. So we need an appropriate metric to measure the distortions of the values of the features for measure the robustness of the features. The scales of the values of the features should not influence measured distortions, otherwise comparing distortions will become meaningless and, more importantly, encouraging distortions to be small will make the scales of the values of the features become small. Therefore, we apply batch normalization \cite{ref5} to all networks in this paper and we only measure the distortions of the normalized values of features.

For a given feature, we measure the distortion d between the original value of it and the distorted value of it as $d=(z-z^{*})^{2}$, where z is the normalized value of the feature and $z^{*}$ is the normalized distorted value of the feature. In testing phase, we compute the mean distortions of the values of the features in a layer to evaluate the robustness of the features in that layer.

Now we perform an experiment to evaluate the robustness of the features learned by standard neural networks and adversarially trained neural networks. Because FGSM cannot find adversarial examples reliably, especially for adversarially trained (against FGSM) networks \cite{ref9}, we do not use the examples generated by FGSM to evaluate the robustness of the learned features. And because PGD is a universal first-order adversary \cite{ref4}, i.e. the strongest adversary utilizing the local first order information, we use PGD to generate adversarial examples to evaluate the robustness of the learned features.

We first use normal technique and adversarial training (against FGSM) to train 2 networks on MNIST respectively and 2 networks on CIFAR-10 respectively (the network architectures can be found in Appendix). Then we compute the mean distortions of the values of the features in every \emph{normalization layer} caused by adversarial perturbations, as is shown in Fig. 1. We also report the accuracies that the 4 networks achieve on different test data in Table 1.

Although the metric we use to measure the distortions of value of the features is different from the metric used in \cite{ref8}, we also observe the "error amplification effect". And the amplitudes of the distortions indicate that the learned features are not robust. 
We observe that adversarial training increases the robustness of the learned features and makes the trained networks slightly more resistant to PGD attack, which indicates that more robust features lead to higher resistance to adversarial attack. But adversarial training does not make the learned features very robust.
\begin{figure}[htbp]
	\centering
	\subfigure[MNIST]{
		\label{fig:subfig1}
		\includegraphics[width=.35\textwidth]{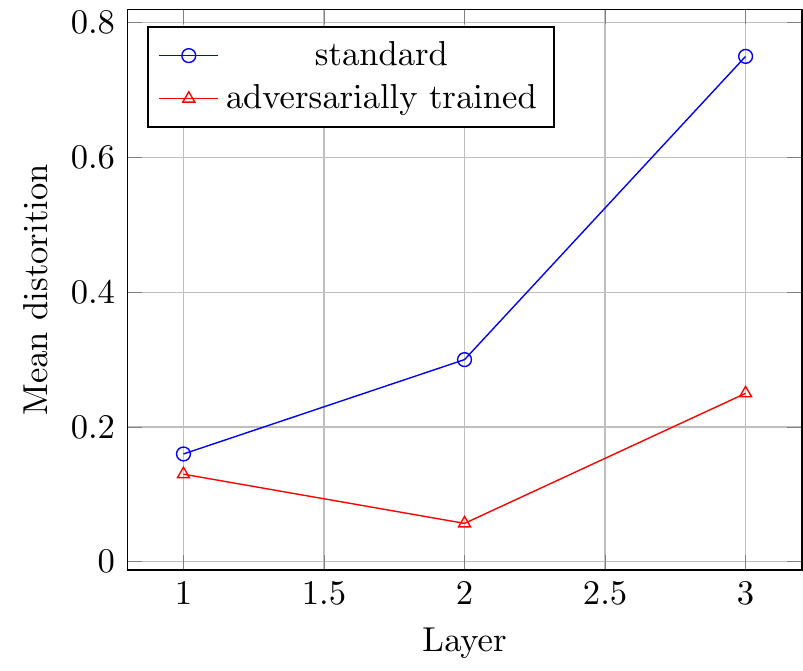}}
	\hspace{1in}
	\subfigure[CIFAR-10]{
		\label{fig:subfig2}
		\includegraphics[width=.35\textwidth]{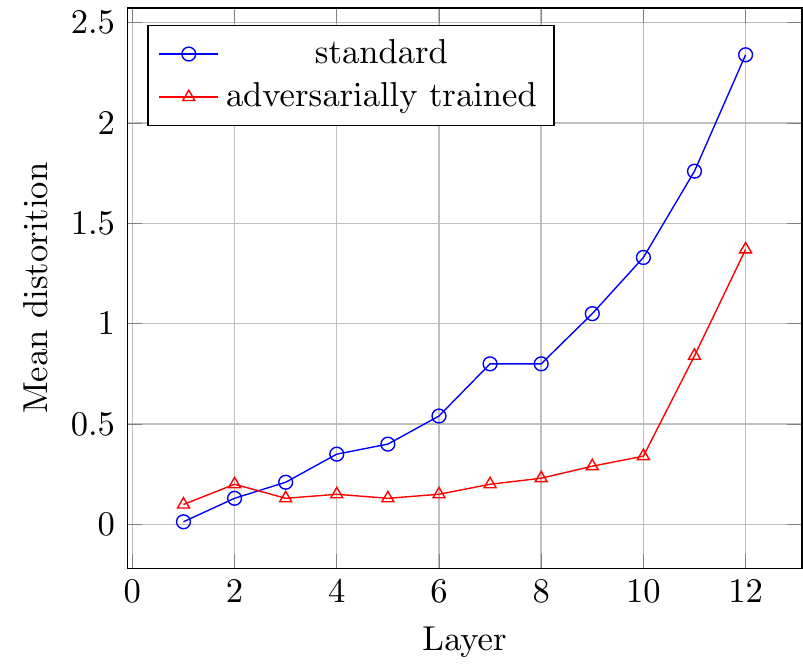}}
	\caption{The mean distortions of the values of the features learned by the 6 networks}
	\label{fig:label1}
\end{figure}
\begin{table}[htbp]
	\caption{The accuracies that the 4 networks (standard two and adversarially trained two) achieve on different test data sets}
	\label{tab:1}
	\centering
	\begin{tabular}{*{7}{c}}
		\toprule
		\multirow{2.7}{*}{\centering Models}
		& \multicolumn{3}{c}{MNIST} 
		& \multicolumn{3}{c}{CIFAR-10} \\
		\cmidrule{2-7}
		\multicolumn{1}{c}{}  
		&\multicolumn{1}{c}{Clean} & {FGSM} & {PGD} & {Clean} & {FGSM} & {PGD} \\
		\midrule
		Standard & 0.9939 & 0.0922 & 0 & 0.9306 & 0.5524 & 0.0256 \\
		\midrule
		Adversarially trained & 0.9932 & 0.9492 & 0.0612 & 0.8755 & 0.8526 & 0.1043 \\
		\bottomrule
	\end{tabular}
\end{table}
\subsection{Proposed Method}
As observed above, more robust features will lead to higher resistance to PGD attack, which means higher resistance to all first-order adversaries because PGD is the strongest first-order adversary \cite{ref4}. In order to make the features learned by neural networks more robust, we add a distortion term to the original adversarial objective function to encourage the distortions to be smaller during training. Formally, we train neural networks with this objective function:
\begin{equation}
\tilde{J}_{adv}(\theta, x, y) = \alpha \cdot J(\theta, x, y) + (1-\alpha) \cdot J(\theta, x^{*}, y) + \sum_{i=1}^{L-1}\beta_i \cdot \sum_{j=1}^{h_i}d_{ij}
\end{equation}
where L - 1 is the number of the normalization layers of the trained network, $h_i$ is the number of features in the ith layer, $d_{ij}$ is the distortion of the value of jth normalized feature in the ith layer, $\beta_i$ is a constant for balancing every term. The adversary for adversarial training can be any adversary, e.g. the universal first-order adversary PGD attack, however we use FGSM as our adversary in this paper for computationally efficiency. According to our experience, the values of $\alpha$ and $\beta_{i}s$ can significantly influence the performance of the trained model, and we find that setting larger $\beta_{i}s$ for higher layers is relatively better.
\section{Experiment}
In this section we evaluate our method on MNIST and CIFAR-10. We evaluate both the resistance against PGD attack and FGSM attack and the robustness of the learned features. The hyper-parameters for training the networks can be found in Appendix. For MNIST, the adversarial examples are generated by PGD with 20 steps of size 0.01($\ell_\infty$) and FGSM with $\epsilon=0.2$. For CIFAR-10, the adversarial examples are generated by PGD with 12 steps of size 1($\ell_\infty$) and FGSM with $\epsilon=4$.
	
The accuracies the trained networks achieve on clean test data and perturbed test data are in Table 2, and the mean distortions of the values of the features learned by the 6 networks are in Fig. 2. We can observe that our method significantly increase the robustness of the learned features and the resistance to both PGD attack and FGSM attack.
	
Despite the significant increase of resistance and robustness, the networks are still not very resistant to adversarial attacks. We believe that using PGD as the adversary during training can make the trained networks more resistant to adversarial attacks and make the learned features more robust.
	
\begin{table}[htbp]
	\caption{The accuracies the trained networks achieve on clean test data and perturbed test data}
	\label{tab:2}
	\centering
	\begin{tabular}{*{7}{c}}
		\toprule
		\multirow{2.7}{*}{\centering Models}
		& \multicolumn{3}{c}{MNIST} 
		& \multicolumn{3}{c}{CIFAR-10} \\
		\cmidrule{2-7}
		\multicolumn{1}{c}{}  
		&\multicolumn{1}{c}{Clean} & {FGSM} & {PGD} & {Clean} & {FGSM} & {PGD} \\
		\midrule
		Standard & 0.9939 & 0.0922 & 0 & 0.9306 & 0.5524 & 0.0256 \\
		\midrule
		Adversarially trained & 0.9932 & 0.9492 & 0.0612 & 0.8755 & 0.8526 & 0.1043 \\
		\midrule
		Our method & 0.9903 & 0.9713 & 0.9171 & 0.8714 & 0.6514 & 0.3440 \\
		\bottomrule
	\end{tabular}
\end{table}
\begin{figure}[htbp]
	\centering
	\subfigure[MNIST]{
		\label{fig:subfig3}
		\includegraphics[width=.4\textwidth]{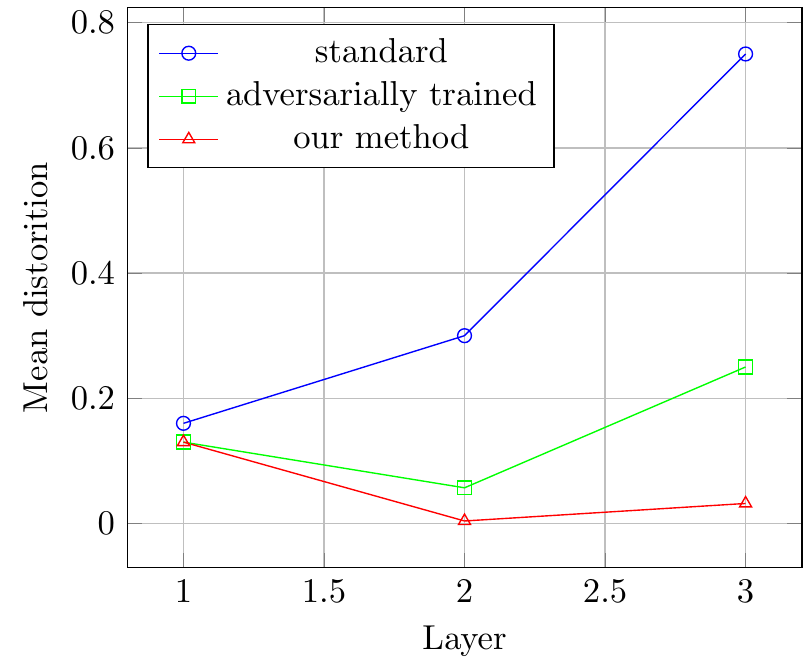}}
	\subfigure[CIFAR-10]{
		\label{fig:subfig4}
		\includegraphics[width=.4\textwidth]{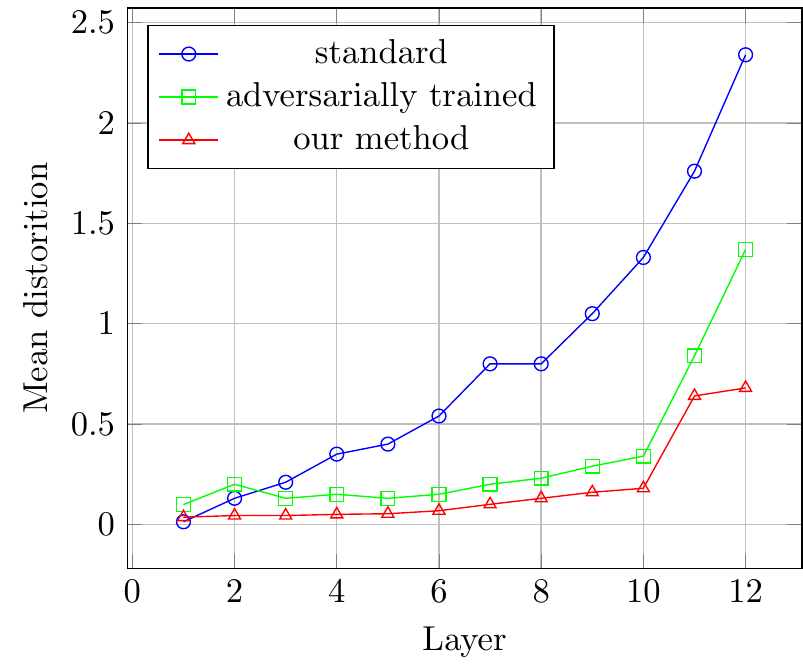}}
	\caption{The mean distortions of the values of the features learned by the 6 networks. Note that the distortions are caused by PGD attack due to the reason mentioned in Section 3}
	\label{fig:label2}
\end{figure}
\section{Conclusion}
In this short paper, we studied the robustness of the feature representations learned by the neural networks trained by standard technique and adversarial training and, based on the observations, we proposed a simple method to train networks to learn more robust features and thus to be more resistant to the adversarial attacks.
	
Our method is similar to the method proposed in \cite{ref7} in the sense that they both leveraging adversarial training to force the distortions of internal features or representations caused by adversarial perturbations to be small. But their method only leverages the representation in one layer of a network and the representation is not normalized and, most importantly, their method has not been shown to be able improve the resistance to some stronger adversaries such as PGD attack.

\section*{Appendix}
We detail the settings of the neural networks used in our experiments here.
	
The architectures of the neural networks for MNIST and CIFAR-10 are in Table 3. Note that batch normalization is applied to the networks. The hyper-parameters for training are in Table 4.
\begin{table}[htbp]
	\centering
	\caption{The architectures of the neural networks for MNIST and CIFAR-10}
	\label{tab:3}
	\begin{tabular}{ll}
		\toprule
		MNIST & CIFAR-10 \\
		\midrule
		Conv(5, 5, 1, 64) + ReLU & Conv(3, 3, 3, 64) + ReLU \\
		Maxpool(2, 2) & Conv(3, 3, 64, 64) + ReLU \\
		FC(1024) + ReLU & Conv(3, 3, 128, 128)\\
		Conv(5, 5, 64, 128) + ReLU & Maxpool(2, 2) \\
		Maxpool(2, 2) & Conv(3, 3, 64, 128) + ReLU \\
		FC(1024) + ReLU & Conv(3, 3, 128, 128)\\
		Softmax(10) & Maxpool(2, 2) \\
		& Conv(3, 3, 128, 256) + ReLU \\
		& Conv(3, 3, 256, 256) + ReLU \\
		& Conv(3, 3, 256, 256) + ReLU \\
		& Maxpool(2, 2) \\
		& Conv(3, 3, 512, 512) + ReLU \\
		& Conv(3, 3, 512, 512) + ReLU \\
		& Conv(3, 3, 512, 512) + ReLU \\
		& Maxpool(2, 2) \\
		& FC(1024) + ReLU \\
		& FC(1024) + ReLU \\
		& Softmax(10) \\
		\bottomrule
	\end{tabular}
\end{table}
\begin{table}[htbp]
	\centering
	\caption{The hyper parameters for training}
	\label{tab:4}
	\begin{tabular}{*{18}{c}}
		\toprule
		\multicolumn{1}{c}{Training method} & & \multicolumn{1}{c}{$\alpha$} & \multicolumn{1}{c}{$\epsilon$} & $\beta_{i}s$ \\
		\midrule
		\multirow{2.5}{*}{\centering Adv. training}
		& MNIST & 0.2 & 0.2 & $\times$ \\
		\cmidrule{2-5}
		\multicolumn{1}{c}{}  
		& CIFAR-10 & 0.2 & 4 & $\times$ \\
		\midrule
		\multirow{2.2}{*}{\centering Our method}
		& MNIST & 0.2 & 0.2 & 1e-7, 1e-7, 3e-7 \\
		\cmidrule{2-5}
		& CIFAR-10 & 0.2 & 4 & \minitab[c]{0, 0, 3e-8, 9e-8, 12e-8, 15e-8, \\ 18e-8, 21e-8, 24e-8, 48e-8, 48e-8} \\
		\bottomrule
	\end{tabular}
\end{table}
	
\end{document}